\documentclass{llncs}
\usepackage{mathptmx}
\usepackage{url}
\usepackage[numbers,sort]{natbib}   
\usepackage{dashbox}
\usepackage{ccg-latex}
\usepackage{expex}
\usepackage{stmaryrd}
\newcommand{\xref}[1]{\,(\ref{#1})}
\newcommand{\xreff}[2]{\,(\ref{#1}#2)}
\lingset{
    belowglpreambleskip=-0.2ex,
    everygla=,
    aboveglftskip=-0.2ex,
    interpartskip=0ex,
    glspace=!0pt plus .2em,
    glrightskip=0pt plus .5\hsize,
    aboveexskip= .4ex plus .4ex minus .4ex,
    belowexskip=.4ex plus .4ex minus .4ex
}
\begin{document}
\title{Paracompositionality, MWEs\\ and Argument Substitution}
\titlerunning{Paracompositionality, MWEs and Argument Substitution}  
%
\author{Cem Boz\c{s}ahin  \and Arzu Burcu G\"uven}
\authorrunning{Boz\c{s}ahin and G\"uven}
\institute{Cognitive Science Department, Informatics Institute\\
 Middle East Technical University (ODT\"U), Ankara Turkey\\
 \email{bozsahin@metu.edu.tr}~~ \email{arzuburcuguven@gmail.com}
 }

\maketitle              

\begin{abstract}
Multi-word expressions, verb-particle constructions, idiomatically combining phrases, and phrasal idioms have something in common: not all of their elements contribute to the argument structure of the predicate implicated by the expression.\smallskip

Radically lexicalized theories of grammar that avoid string-, term-, logical form-, and tree-writing, and categorial grammars that avoid wrap operation, make predictions about the categories involved in verb-particles and phrasal idioms. They may require singleton types, which can only substitute for one value, not just for one kind of value. These types are asymmetric: they can be arguments only.
They also narrowly constrain the kind of semantic value that can correspond to such syntactic categories.
Idiomatically combining phrases do not subcategorize for singleton types, and they
exploit another locally computable and compositional property of a correspondence, that every syntactic expression can project its head word.
Such MWEs can be seen as empirically realized categorial possibilities
 rather than 
lacuna in a theory of lexicalizable syntactic categories.\\

\keywords{Syntax, semantics, CCG, multi-word expression, idiom, verb-particle, lexical insertion, type theory}
\end{abstract}

\section{Introduction}
A type is a set of values. When we write a syntactic type, say {NP}, we mean a set of expressions 
(values) which can substitute for that type. This type serves to distinguish some expressions from for example the
set of expressions that can substitute for a {VP} type.

The distinction is crucial for solving the correspondence problem in syntax-semantics. For this purpose we talk about semantic types, for example $e$ for things and $t$ for propositions. The concepts that can substitute for semantic types  are not expressions in the sense that syntactic expressions are, because they are not observable, but they leverage a theory to hypothesize about the kind of semantics that these types stand for.

These two species of types are then put in a correspondence in a theory of syntax-semantics connection.
The understanding is that if one substitutes a certain expression for a syntactic type, then its corresponding semantic type substitutes for a certain kind of semantic value. We know less about the semantic values; but, at the level of the correspondence problem, this is not very critical. It is however crucial to make the distinctions and propagate them in a parsing mechanism rather than solving all type-interpretation problems in one go.

We need a theory which provides explicit vocabulary and mechanism for the correspondence, to be more specific about the equal relevance of substitution for subexpressions which purportedly do not contribute to the meaning
of the expression.

In the categorial grammar parlance, for which we will use Combinatory Categorial Grammar \cite{Stee:99,Stee:11a}, hereafter CCG, we can exemplify the correspondence as follows, where we use
the ``result-first argument-next'' notation:

\pex\label{ex:hits}
\a hits := \cgf{(S\bs\cgs{NP}{3s})\fs NP}$:\lambda x\lambda y.\so{hit\,}xy$
\a hit := \cgf{\cgs{VP}{inf}\fs NP}$:\lambda x\lambda y.\so{hit\,}xy$
\xe
Some syntactic types are further narrowed down by features, such as \cgs{NP}{3s} above for third person singular NP,
which are, in CCG, not re-entrant.

We argue in the paper that in a radically lexicalized theory
which adheres to transparency of derivations by type substitution (rather than lexical insertion), such as CCG, 
there are built-in degrees of freedom  to support Multi-word Expressions (MWEs) and idioms
without complicating the mechanism. 

Paracompositionality is key to projection of their properties in a derivation. It is the idea that,
in addition to the compositionality of the lexical correspondence, which is compositional partly because it relies on non-vacuous abstractions, type substitution by (i) what we call singleton types and (ii) what is called
head-dependencies in the NLP literature
is also compositional because it spells non-vacuous abstraction as part of the correspondence, but as something related to the contingency of the predicate, rather than
the argument structure of the predicate. In a radically lexicalized grammar
both sources are available in a lexical item. These types are paracompositional also in the sense that whether we have an idiom reading
or compositional one is already decided by the category of the head 
in the derivational process.

The term \emph{contingency} is used here in the sense of Moens and Steedman \cite{Moen:88a} 
where it relates to extension of happenings. In the case of events (culminations, points, processes and culminating processes),
which have definite extension, it is an event modality of space, time and manner; and, in the case of states where extension is indefinite (e.g. \emph{understand}) it is
some property of the state. From now on when we use the term `contingency' we mean something related to
extension of the predicate, rather than who does what to whom in the predicate.

MWEs are expressions involving 
more than one word in which the properties of the expression are not determined by the composition of the properties of the constituent words, which would be the case for phrases. There is a tendency to treat them as single lexical units \cite{cope:94,vill:04}; but, as we shall see, CCG does not require the single unit to be the 
phonological representation to the left of `:=' in the format of\xref{ex:hits}. This property of CCG naturally extends to coverage of verb-particle constructions
e.g. \emph{look the word up}
as discontiguous MWEs headed by a lexical item. 

Phrasal idioms and idiomatically combining phrases are
classes identified by Nunberg, Sag and Wasow \cite{nunb:94} to account for systematic variation in syntactic productivity of idioms.
Typewise they will relate to singleton types (phrasal idioms) and head-word subcategorization (idiomatically combining phrases) in our formulation.

As a preview of the article, we can think of the meaning distinctions as ranging from ``beans''
i.e. the nounphrase \emph{beans} itself as a category (this is what we call the
singleton type);
to
\cgs{NP}{beans} as the category of an NP headed by the word \emph{beans}, which has wider range of substitution; and, to the polyvalent NP with the widest substitution for that type. This much is categorial grammar with type substitution. CCG as an empirical theory adds to this the claim that
there is an asymmetry in the range of substitutions: the singleton types
can be arguments only, and arguments of arguments and results, but never the result.
We shall see that this has implications for the linguist's choice of handling syntactic productivity in a grammar.

Some implications follow: Because of paracompositionality,
all expressions requiring a singleton type
would involve the semantic type of a predicate,  and all idiomatically combining phrases requiring
a different interpretation than the compositional one would have the same consequence independent of their syntactic productivity. In short, every idiom must contain a predicate (but not necessarily a verb).
We cover these
implications in the article.

\section{Substitution in a Derivation}

In\xreff{ex:hits}{a}, the `\cgf{\fs NP}' can be substituted for by certain kinds of expressions, for example \emph{John,
me, the ball, a stone in the corner}, etc. Its corresponding semantic counterpart in the logical form (LF), written after the colon, has the placeholder $x$ which can be typed as $e$, to be suitably substituted for by a semantic value described above. The `\cgf{\bs\cgs{NP}{3s}}' can be substituted by narrower expressions, for example eliminating  \emph{I, you}. Because this is an indirect correspondence, its semantic counterpart $y$ can have the same type $e$. 

The tacit assumption of indirectness is sometimes made explicit,
for example in  Bach's \cite{Bach:76} rule-by-rule hypothesis: The derivational process operates with syntactic types only, and when it applies the semantics of the rule,  its semantics works only with LF objects. Quoting from Bach: ``Neither type of rule has access to the representations of the other type except at the point where a translation rule corresponding to a given syntactic rule is applied.'' The ``syntactic rule'' in a lexicalized grammar such as CCG is the combinatory syntactic type of a lexical correspondence. The`` translation rule'' is the lexically-specified logical form, LF, as in\xref{ex:hits}. 

The derivational process reveals partially derived types, for example \cgf{S\bs\cgs{NP}{3s}}$:\lambda y.\so{hit\,}\so{s}y$ for\xreff{ex:hits}{a}, if function application substitutes say \emph{a stone} for `\cgf{\fs NP}', with
some semantic value \so{s}. The semantic type of such derived categories is concomitantly functional,
e.g. $e\mapsto t$ for this syntactic type. \emph{John hits} is $e\mapsto t$ too, with category \cat{S\fs NP}\lf{\lambda x.\so{hit\,}x\,\so{john}}.

We can see the relevance of derived types to substitutability in a closer look at\xreff{ex:hits}{b}.
If function application substitutes for the `\cgf{\fs NP}' in the example, the derived category would be
\cgf{\cgs{VP}{inf}}$:\lambda y.\so{hit\,}\so{s}y$\, in this case. This is also an $e\mapsto t$ type semantically. However,
its syntax is narrower so that we can account for the expressions in\xref{ex:badvp}.\footnote{This is equivalent to
saying that in CCG the type \cgf{VP} is not always an abbrevation for \cgf{S\bs NP}, which might be the case
in other brands of categorial grammars. The English facts above could be taken care of
by featural distinctions such as \cat{\cgs{S}{inf}, \cgs{S}{to-inf}, \cgs{S}{fin}} in \cat{S\bs NP}, rather than also positing a \cat{VP}.
But in  ergative languages  the \cat{`\bs NP'} does not always coincide with the same LF role as it does in English,
such as in Dyirbal's control construction, where the controlled absolutive argument can be the patient NP of the transitive clause or syntactic subject of an intransitive clause, but not the ergative NP of the 
transitive clause. It seems to require
\cat{VP}\lf{\lambda x.\so{pred\,}x} where $x$'s role in the controlled clause \so{pred\,} is determined
by verbal morphology of the controlled clause; see \cite{manning96} for the phenomenon. Assuming a \cat{VP} cross-linguistically makes narrower predictions about control.
We handle this problem elsewhere.}

\ex\label{ex:badvp}
John persuaded Mary to/*~~hit/*hits the target.
\xe

The derivational process works  as below, with \cgs{VP}{to-inf} distinct from \cgs{VP}{inf}.
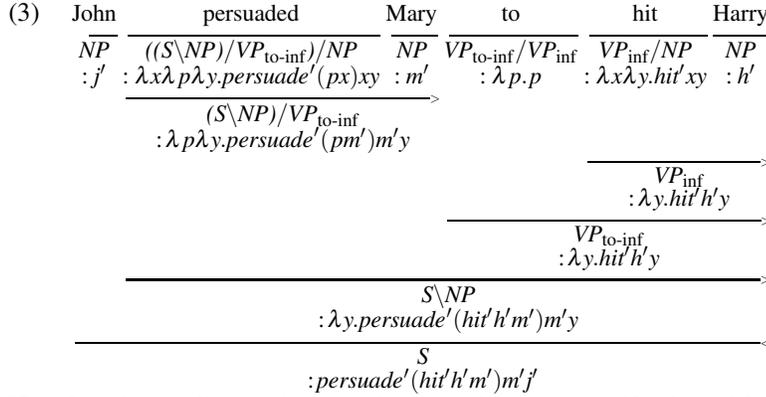
\ex\small
\begin{ccg}{6}{John & persuaded & Mary & to & hit &Harry}
{
\cgf{NP} 
& \cgf{((S\bs NP)\fs\cgs{VP}{to-inf})\fs NP}
& \cgf{NP}
& \cgf{\cgs{VP}{to-inf}\fs\cgs{VP}{inf}}
& \cgf{\cgs{VP}{inf}\fs NP}
& \cgf{NP}\\
$:\so{j}$ 
&$:\lambda x\lambda p\lambda y.\so{persuade}(px)xy$
&$:\so{m}$
&$:\lambda p.p$
& $:\lambda x\lambda y.\so{hit}xy$
&$:\so{h}$\\
& \cgline{2}{\cgfa}\\ &\cgres{2}{(S\bs NP)\fs\cgs{VP}{to-inf}}\\
&\cgres{2}{\lf{\lambda p\lambda y.\so{persuade}(p\so{m})\so{m}y}}\\
&&&& \cgline{2}{\cgfa}\\ &&&&\cgres{2}{\cat{\cgs{VP}{inf}}}\\
&&&&\cgres{2}{\lf{\lambda y.\so{hit}\so{h}y}} \\
&&&\cgline{3}{\cgfa}\\ &&&\cgres{3}{\cgs{VP}{to-inf}}\\
&&&\cgres{3}{\lf{\lambda y.\so{hit}\so{h}y}}\\
&\cgline{5}{\cgfa}\\ &\cgres{5}{\cat{S\bs NP}}\\
& \cgres{5}{\lf{\lambda y.\so{persuade}(\so{hit}\so{h}\so{m})\so{m}y}}\\
\cgline{6}{\cgba}\\ \cgres{6}{S}\\
\cgres{6}{\lf{\so{persuade}(\so{hit}\so{h}\so{m})\so{m}\so{j}}}
}

\end{ccg}
\xe
 Here function application is shown in forward
form (\cgfa) and backward form (\cgba). Derivation proceeds from top to bottom in display, as standard in CCG;
i.e., bottom-up as far as parsing is concerned,
and one at a time. For brevity alternative derivations using function composition are not shown; their
implications for constituency are discussed in Steedman references. We also eschew the slash modalities
of Baldridge and Kruijff \cite{Bald:03} to avoid digression, which can further restrict the combination possibilities
of syntactic types. They are mentioned later when they are relevant to discussion.
The LF contains a structured form, viz. the predicate-argument structure, which is written in linear notation for simplicity; for example
$\so{hit}xy$ is same as $((\so{hit}x)y)$; i.e., it is left-associative.

\begin{figure}
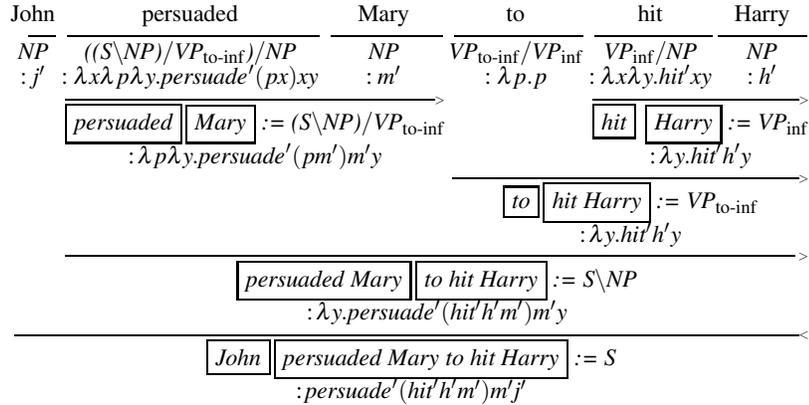

\begin{ccg}{6}{John & persuaded & Mary & to & hit &Harry}
{
\cgf{NP} 
& \cgf{((S\bs NP)\fs\cgs{VP}{to-inf})\fs NP}
& \cgf{NP}
& \cgf{\cgs{VP}{to-inf}\fs\cgs{VP}{inf}}
& \cgf{\cgs{VP}{inf}\fs NP}
& \cgf{NP}\\
$:\so{j}$ 
&$:\lambda x\lambda p\lambda y.\so{persuade}(px)xy$
&$:\so{m}$
&$:\lambda p.p$
& $:\lambda x\lambda y.\so{hit}xy$
&$:\so{h}$\\
& \cgline{2}{\cgfa}&& \cgline{2}{\cgfa}\\ 
&\cgres{2}{\fbox{persuaded}\,\fbox{Mary} := (S\bs NP)\fs\cgs{VP}{to-inf}}&&\cgres{2}{\fbox{hit}\, \fbox{Harry} := \cat{\cgs{VP}{inf}}}\\
&\cgres{2}{\lf{\lambda p\lambda y.\so{persuade}(p\so{m})\so{m}y}}&&
\cgres{2}{\lf{\lambda y.\so{hit}\so{h}y}} \\
&&&\cgline{3}{\cgfa}\\ &&&\cgres{3}{\fbox{to}\,\fbox{hit Harry} := \cgs{VP}{to-inf}}\\
&&&\cgres{3}{\lf{\lambda y.\so{hit}\so{h}y}}\\
&\cgline{5}{ \cgfa}\\ &\cgres{5}{\fbox{persuaded Mary}\,\fbox{to hit Harry} := \cat{S\bs NP}}\\
& \cgres{5}{\lf{\lambda y.\so{persuade}(\so{hit}\so{h}\so{m})\so{m}y}}\\
\cgline{6}{\cgba}\\ \cgres{6}{\fbox{John}\,\fbox{persuaded Mary to hit Harry} := S}\\
\cgres{6}{\lf{\so{persuade}(\so{hit}\so{h}\so{m})\so{m}\so{j}}}
}

\end{ccg}
\caption{Substitution of syntactic expressions for syntactic types. Boxes show segments combined.
We display some one-at-a-time derivations
on the same line to save space.
}
\label{fig:subs}
\end{figure}

In preparation for final discussion of substitution (\S\ref{sec:nowrap})
in relation to the wrapping operation, we can redraw this derivation
by showing the substituting expressions as we proceed, which we do 
in Figure~\ref{fig:subs}.

MWEs present a challenge for substitution  in such correspondences.
In Schuler and Joshi's \cite{schu:11a}:25 words: ``In the \emph{pick .. up} example, there is no coherent meaning for $\mathit{Up}$ such that
$\llbracket\mathit{pick}\,\,X\,\mathit{Up}\rrbracket=\mathit{Pick}(\llbracket X\rrbracket, \mathit{Up})$.'' They go on to show how tree-write in the form of TAG transformations, rather than string-rewrite of CFG transformations such as \cite{sag:02}, can deliver different meanings of such expressions \emph{after}
a fully compositional tree is established for `pick', `..' and `up'.

In such systems, post-processing and reanalysis of a categorial surface derivation are possible, both for TAG and HPSG,\footnote{TAG transformations 
take a phrase structure tree and decompose it to elementary structures
to deliver an LF. \cite{lich:16} is a different TAG way to incorporate meaning postulates of
\cite{pulm:93}.
HPSG uses phrasal post-classification to the same effect. For example
\cite{bond:15,sag:02} perform it at the final stage of parsing as a semantic check on bags of predicates
for idiom entries, and \cite{kay:idioms} use semantic frame identification, viz. compositional vs idiomatic,
which are built in to theory. The diversity of approaches in the volume for idioms \cite{ever:95} 
is testimony to the practice that the idioms are decisive factors in polishing our theories
linguistically, psychologically and computationally.}
 therefore these transformations are possible, indeed useful, to simplify large-scale grammar development.

For radically lexicalized grammars such as CCG where such options are not available, three paths 
to maintaining compositionality in the presence of ``non-compositional'' and/or idiomatic parts seem to be available:
\pex\label{rul:options}
\a letting the logical form change the compositional meaning,
\a introducing surface wrap, 
\a reassessing the substitutability of \emph{argument} types, to the extent
that (i) they can be narrowed by head-dependencies, and
(ii) the semantic contribution of some parts of the correspondence to the predicate-argument structure
can be ignored in a principled way, and locally.
\xe

The problem is exacerbated by phrasal idioms which seem to have partially active syntax in some non-compositional parts, for example \emph{kick the (proverbial/old) bucket}, 
but note $\sharp$\emph{the bucket that John kicked}, $\sharp$\emph{kick the great bucket in the sky}, and *\emph{the breeze was shot}. ($\sharp$ is used to indicate unavailability of idiomatic reading. The last two examples and judgments are from \cite{sag:02}.) However, there are also phrasal idioms which are  syntactically quite active, e.g. \emph{the beans that John spilled}, and \emph{spilling the musical/artistic/juicy beans}.

Option\xreff{rul:options}{a}
does not always necessitate post-processing of  MWEs in CCG, but, as we shall see later in\xref{ex:posthoc}, it does not guarantee locality of derivations either. One way to realize it is the following:
\ex \label{ex:kick0}
kicked := \cat{(S\bs NP)\fs NP}
\lf{\lambda x\lambda y.\mathrm{if\,} {head}(x)=\so{bucket\,} \mathrm
{then\,} \so{die}y\,\mathrm{else\,} \so{kick}xy}
\xe

This approach to phrasal idioms which is similar to meaning postulates for the same task such as
\cite{pulm:93} 
would then have to make sure that the head meaning \so{bucket\,} has some predefined
cluster of modifiers such as \emph{proverbial} or \emph{old}, but not much else, for example \emph{$\sharp$kick the bucket that overflowed}. It would also have to overextend itself to avoid the idiomatic reading
in \emph{$\sharp$the bucket that you kicked}. 

As an alternative, the type \cgs{NP}{bucket} below is inspired by trainable stochastic CFGs which
can distinguish argument PPs from adjunct PPs by encoding head dependencies
for CFG rules, for example VP$_{\mathrm{put}} \rightarrow$ V$_{\mathrm{put}}$~NP~PP$_{\mathrm{on}}$:
(We shall fix the unaccounted vacuous abstraction
in it later in the paper.)

\ex\label{ex:kick0-b}
kicked := \cat{(S\bs NP)\fs\cgs{NP}{bucket}}\lf{\lambda x\lambda y.\so{die}y}
\xe

It might appear to be LF-motivated just like\xref{ex:kick0} above; but, it is actually a case of\xreff{rul:options}{c/i}.
\cgs{NP}{bucket}, meaning NP headed by \emph{bucket}, can  be made distinct from \cgs{NP}{buckets} because
different surface expressions can be substituted for them.\xref{ex:kick0-b} overgenerates for the examples given above, but it might be the right degree of freedom to exploit in the syntax-semantics correspondence of
idiomatically combining MWEs such as 
\cgs{NP}{beans} for \emph{spill the beans}.

In the remainder of the paper, we show that option\xreff{rul:options}{c/ii} has been implicit in CCG theory all along but never used, in the form of syntactic types for which only one value can substitute (\S\ref{sec:singletons}). 
We call them singleton types.
This way of lexical categorization and subcategorization predicts very limited syntax, but not as metalinguistic marking that \cite{sag:02} proposed for \emph{kick the proverbial/old bucket}. It is due to having to enumerate different senses and contingencies of phrasal idioms (e.g. proverb bucket for senses above, also covering
e.g. \emph{when I face the proverbial bucket}), and \emph{pick up} for MWEs. In \S\ref{sec:icp} we show that idiomatically combining phrases
have principled distinctions from singleton types.
Head-word subcategorization such 
as\xref{ex:kick0-b} is the more promising option for them, which radically lexicalized grammars can handle without extension. There are also idioms which require analysis combining both options such as those
with semantic reflexives where the referent is not part of the idiom, e.g. \emph{I twiddle my/*his thumbs}.
\S\ref{sec:combined} covers these cases.

These  findings reveal some aspects of type substitution and its projection when the expressions are not fully compositional at the level of the predicate-argument structure.
As such they may have implications beyond CCG.

Finally we show that adopting option\xreff{rul:options}{b} 
to analyze for example \emph{pick $\cdots$ up} as \emph{pick up $(\cdots)_{\mathrm{wrap}}$}
overgenerates in the combinatory version of wrap  (\S\ref{sec:nowrap}), and complicates the grammar with a domino-effect in the surface version of wrap; therefore, it would do more damage than good
if adopted for (discontiguous) MWEs and phrasal idioms. CCG can continue to avoid all forms of wrap in the presence of all kinds of MWEs and phrasal idioms.

\section{Singleton Types}\label{sec:singletons}

A brief preview of the proposal for\xreff{rul:options}{c/ii} is as follows. A singleton syntactic type self-represents because it can substitute for  one value only. We designate such types with strings, such as \cgf{``up''} or \cgf{``the bucket''}; for example:
\pex\label{ex:single1}
\a picked := \cat{(S\bs NP)\fs``up''\fs NP}\lf{\lambda y\lambda x\lambda z.\so{cause}(\so{init}(\so{hold$_x$}yz))z}
\a kicked := \cat{(S\bs NP)\fs``the bucket''}\lf{\lambda x\lambda y.\so{die$_x$}y}
\xe
(\so{Init} is a function that yields
a culminating state in the sense of \cite{Moen:88a}.)

We call categories in\xref{ex:single1} `paracompositional' to highlight
the fact that, although their LF correspondence is intact so that the derivational process is transparent, they might have seemingly vacuous abstraction from the
perspective of the predicate-argument structure, symbolized by the placeholders $x$ above.\footnote{van der Linden \cite{lind:92}, which is another categorial approach to idioms, allows vacuous abstractions, i.e. define semantics without mention of $x$ in the LF of\xreff{ex:single1}{b}.
Apart from our empirical claim that they have a place in LF because they relate to contingency,
vacuous abstractions seem to open ways to resource insensitivity which is unheard of in natural language; for example, the \combk\,combinator with its vacuous abstraction $\lambda x\lambda y.x$ can delete things from LF.
We have yet to find a word or morpheme that does this; see \cite{bozs:12}:81 for some speculation.

\cite{lind:92}'s treatment of phrasal idioms such as \emph{kick the bucket} assumes partial
involvement of the head verb \emph{kick} for the semantics of the idiom, whereas in our conception it is fully responsible
for the idiom with the aid of singleton types.}

However, one can make a case that this abstraction, corresponding
respectively to singleton categories \cat{``up''} and \cat{``the bucket''}, might have a role
inside the LF constants shown in primes, as contingencies.
We write them for example as $\so{die}_xy$ (as ceremonial death, reported death, etc.),
rather than $\so{die}y$.
These LF `constants' are convenient generalizations in CCG standing in for a plethora of features anyway, so it seems natural to think of them as having
their own abstraction. (The semantic types
corresponding to these contingencies are then $\alpha\mapsto t$ for some $\alpha$.)

It will be seen in \S\ref{sec:analysis} that the examples in\xref{ex:single1} differ in their sense
from \emph{picked up the book} and \emph{kicked the blue bucket}, therefore a separate grammar entry is empirically justified.
The sense distinction is reflected explicitly in the LF, as we shall see later. Both possibilities for substitution, for the syntactic type and for its placeholder in the LF, are principally restricted by CCG.

Singletons also engender a way for such entries to be morphologically more transparent, for example by being susceptible to inflection, e.g. \emph{picking}, by providing a segmental alternative to contiguous but MWE \emph{pick up $\cdots$}, which would need a morphological pointer
for inflection, as noted by \cite{sag:02,vill:04} for their analyses. Nunberg, Sag and Wasow's \cite{nunb:94} dichotomy between phrasal idioms and idiomatically combining items also vanishes, because of the singleton types
and head-word subcategorized argument types.
The distinction between
syntactically pseudo-active \emph{kick the bucket} and more active \emph{spill the beans} naturally follows from whether the idiomatic part has a role in the predicate-argument structure, which we capture 
by systematically choosing between option\xreff{rul:options}{c/i} and\xreff{rul:options}{c/ii} per lexical correspondence.

\subsection{Parsing and Correspondence with Singleton Types}
The crucial property of a category in a lexical correspondence such as $\alpha$ := \cat{A\fs``s''} with singleton
\cat{s}, is that the string ``s'' \emph{as a category} does have its own correspondence. This cannot be a literal match without categorial processing of the surface string, with \cat{s} to the right of $\alpha$. It is a compositional derivational process arising from (a) below, to lead to (b). 
The lexically specifiable difference from a polyvalent category such as \cat{NP, VP} is that 
the item $\alpha$ subcategorizes
for the string \cat{s}, hence treat it as a category, rather than subcategorize for the category of \cat{s}, viz. \cat{B} in the example. To obtain \cat{B}, the derivational process works as usual for \emph{s}, independent of the item $\alpha$. We shall see in\xref{rul:fa} that rules of function application need no amendment for this interpretation.\xreff{ex:singletonapp}{b} is lexically determined by $\alpha$.
\pex\label{ex:singletonapp}
\a s := \cat{B}\lf{\so{s}}
\a \begin{ccg}{2}{$\alpha$ & \emph{s}}
{
\cat{A\fs``s''}\lf{\lambda x.p_x} & \cat{B}\lf{\so{s}}\\
\cgline{2}{\cgfa}\\ \cgres{2}{$\alpha$s := \cat{A}\lf{p_{\so{s}}}}
}
\end{ccg}
\xe
Same idea  applies to backward application, for $\alpha$ := \cat{A\bs``s''} and
the sequence $s\alpha$.

In other words, the surface string \emph{s} is derived by the derivational process as well. It is just that the item $\alpha$ carrying the singleton type as an argument decides what to do with its semantics, which we indicated
schematically above as modal contribution to contingency of $p$, as $p_{x}$ of $\alpha$. This is not post-processing of a category in a radically lexicalized grammar, in which all and only head functors decide what to do with
the semantics of their arguments.

It means that, whether an argument type is polyvalent or singleton, there has to be an LF placeholder for it, otherwise the derivational process, which is completely driven by syntactic types in CCG, cannot proceed. 
It can be seen in the basic primitive of CCG, viz. function application:
\ex \label{rul:fa}
\begin{tabular}[t]{llllr}
\cat{X\fs Y}\lf{f} & \cat{Y}\lf{a} &$\rightarrow$& \cat{X}\lf{fa} & 
\hspace*{19em}(\cgfa)\\ 
\cat{Y}\lf{a} & \cat{X\bs Y}\lf{f}  &$\rightarrow$& \cat{X}\lf{fa} & (\cgba)\\
\end{tabular} 
\xe
The LF of the functor, $f$, has to be a lambda abstraction, to be able to take any \cat{Y} and yield $fa$. This is true of singleton `\fs\cat{Y}' and `\bs\cat{Y}' too. 

We can clearly see the role of substitution rather than insertion in projection of types. The rule (\cgfa) above
is in fact realized as below (similarly for others):
\ex\label{rul:fa-expr}
$\alpha$ := \cat{X\fs Y}\lf{f}~~~$\beta$ := \cat{Y}\lf{a}~~~$\rightarrow$~~~$\alpha\beta$ := \cat{X}\lf{fa}  
\hspace*{10em}(\cgfa)\
\xe
There is no sense in which we can insert something into $\alpha$ and $\beta$ as they form $\alpha\beta$ because these are surface expressions.

The singleton types  present an  asymmetry in argument-result (or domain-range) specification.
Functors such as \cat{A\fs B} and \cat{A\bs B} have domain \cat{B} and range \cat{A},
and, apart from trivial identities where \cat{A} and \cat{B} are the same singleton, 
the interpretation where the range itself (\cat{A}) is a singleton is problematic. Since \cat{A\us B}
is a function \emph{into} \cat{A} for some slash `\us', if it is not a trivial case of singleton identity, say \cat{``up''\fs``up''}, it is difficult to see how \cat{A} can be singleton. Although there are no formal reasons
to avoid singleton results, and results of results, we conjecture that singletons are arguments, and arguments of results and arguments, because there seems to be no nontrivial  function of a singleton-result
with grammatical significance.

A related argument can be made about a singleton's potential to be the overall syntactic category of a lexical item. The notion of extending the phonological range of an item such as (a) below coincides naturally with ``words with spaces'' idea (e.g. \emph{ad hoc, by and large, every which way}), which is common in NLP of MWEs, but (b) is also an option. 
\pex \label{ex:eww}
\a every which way := \cat{(S\bs NP)\bs(S\bs NP)}\lf{\lambda p\lambda x.\so{omni\,}px}
\a every which way := \cat{``every which way''}\lf{\so{omniway\,}} 
\xe

Notice that (b) is different than having \emph{scored} := \cat{(S\bs NP)\fs``every which way''} for lexically specified verbal adjunction in the manner of \cite{dowt:03}, which, given\xref{ex:singletonapp}, must either  use entries similar to\xref{ex:eww}, or
derive \emph{every which way} syntactically,  and choose to
trump its category  because it wants a narrower LF due to singleton subcategorization. However we think that both options may be redundant, because of the following.


In CCG  the head functor decides the semantics
of its entry even if it subcategorizes for a singleton category.
Therefore the entries in (a--b) above which we use in (a-b) below \emph{may} be redundant if the words in ``words with spaces'' are part of the grammar, and if they can combine in any way, say as in (c) below for some \cat{A, B, C}:
\pex\label{ex:basicsingleton}
\a \begin{ccg}{3}{My team & scored & every which way} 
{
\cat{NP} & \cat{(S\bs NP)\fs``every which way''} &  \cat{(S\bs NP)\bs(S\bs NP)}\\
&\cgline{2}{\cgfa}\\ &\cgres{2}{\cat{S\bs NP}}
}
\end{ccg}\smallskip

\a \begin{ccg}{2}{scored & every which way}
{
 \cat{(S\bs NP)\fs``every which way''} & \cat{``every which way''}\\
 \cgline{2}{\cgfa}\\ \cgres{2}{S\bs NP}
}
\end{ccg}\smallskip

\a \begin{ccg}{4}{scored & every & which & way}
{
 \cat{(S\bs NP)\fs``every which way''} & \cat{A\fs B}
 & \cat{B\fs C} & \cat{C}\\
 &&\cgline{2}{\cgfa}\\ &&\cgres{2}{B}\\
 & \cgline{3}{\cgfa}\\ & \cgres{3}{A}\\
 \cgline{4}{\cgfa}\\ \cgres{4}{S\bs NP}
}
\end{ccg}
\xe
There would be no post-processing or reanalysis in these cases; they would be multiple analyses because of redundancy. The
transparency of derivation requires that in configurations like\xreff{ex:singletonapp}{b}
the constituents of the rule applying can themselves be derived.

The rules that allow CCG to rise above function application in projection,  composition and substitution also maintain the transparency of the syntactic process, by being
oblivious to the nature of argument types in these rules:\footnote{We show only one directional variant
of each rule for brevity. The same idea applies to all variants; see Steedman references for
a standard set of rules, and \cite{bozs:12} for review of proposals for combinatory extensions.

\noindent Boz\c{s}ahin \cite{bozs:12}:\S 10 shows that all projection rules of CCG can be packed into one monad to enable monadic computation with just one rule of projection. This is possible because CCG is radically lexicalized
in the sense that combinatory rules cannot project anything which is not in the lexicon.
What appears to be rule choice when presented as (\ref{rul:fa}/\ref{rul:cs}) becomes dependency passing
within monad with one rule of combination.
}
\ex \label{rul:cs}
 \begin{tabular}[t]{llllr}
\cat{X\fs Y}\lf{f} & \cat{Y\fs Z}\lf{g} &$\rightarrow$& \cat{X\fs Z}\lf{\lambda x.f(gx)} & 
\hspace*{12.4em}(\cgfc)\\ 
\cat{X\fs Y\fs Z}\lf{f} & \cat{Y\fs Z}\lf{g}  &$\rightarrow$& \cat{X\fs Z}\lf{\lambda x.fx(gx)} & (\cgsf)\\
\end{tabular} 
\xe

If the result categories are not singletons, as we argued, then the rules above never face a case where \cat{Y} is a singleton. This means that, since singletons are arguments,
meaning they bear a slash, say `\cat{\us A}' for some slash `\us' in \{\bs, \fs\},
the slash is inherently application-only, equivalently `\cat{\us$_\star$A}' in \cite{Bald:03} terminology.\footnote{The way this is implemented in many CCG systems including ours
is for example to constrain the slashes as follows:

\begin{tabular}[t]{llllr}
\cat{X\fstars Y}\lf{f} & \cat{Y}\lf{a} &$\rightarrow$& \cat{X}\lf{fa} & 
(\cgfa)\\ 
\cat{X\fds Y}\lf{f} & \cat{Y\fds Z}\lf{g} &$\rightarrow$& \cat{X\fds Z}\lf{\lambda x.f(gx)} & 
\hspace*{17em}(\cgfc)
\end{tabular} 

It is easier to describe slash-modal control from the perspective of syntactic types of expressions accessing 
these rules.
`$\star$-rules' are accessible by all categories, `$\diamond$' and `$\times$'  are compatible
only with themselves, and with the most permissive slash.}

This is corroborated
by examples like below where there is no idiomatic reading:
(We show the derivation for the hypothetical case where singletons would be allowed to compose. Typing the singleton as `\cat{\fs$_\star$``the bucket''}'
eliminates the derivation. The slashes in the paper are harmonic `\bds' or `\fds' unless stated otherwise.)
\ex
\scriptsize
\hspace*{-1.5em}\begin{ccg}{7}
{$\sharp$John  kicked & and & Mary & did & not & kick}
{
\cat{S\fs``the bucket''}&
\cat{(X\bstars X)\fstars X}
&\cat{S\fs(S\bs NP)}&
\cat{(S\bs NP)\fs\cgs{VP}{inf}}&
\cat{\cgs{VP}{inf}\fs\cgs{VP}{inf}}&
\cat{\cgs{VP}{inf}\fs``the bucket''}\\
&& \cgline{2}{\cgfc}\\ &&\cgres{2}{S\fs\cgs{VP}{inf}}\\
&& \cgline{3}{\cgfc}\\ && \cgres{3}{S\fs\cgs{VP}{inf}}\\
&& \cgline{4}{\cgfc}\\ && \cgres{4}{S\fs``the bucket''}\\
\cgline{7}{\&}\\ \cgres{7}{\cat{S\fs``the bucket''}}
}
\end{ccg}\\
\cgex{1}{the bucket\\ \cgul\\[.5ex] \cat{NP} }
\xe

For polyvalent types, one-to-one correspondence of syntactic types and placeholder types is meant to capture the thematic structure in CCG, for example for
\emph{the door opened} versus \emph{someone opened the door}, by having two different (albeit related) correspondences for \emph{open}. 

For a singleton, its functor (and there must be one, since they can only be arguments) decides lexically whether there is a
predicate-argument structural role for the placeholder in the LF, as we see in the distinction of \emph{spill the beans}, where \so{secret\,} is an argument of \so{divulge}, versus \emph{kick the bucket}, where \so{bucket\,} or anything related to it is not an argument of \so{die}.

Therefore, for CCG, MWEs and phrasal idioms are not exceptions that need non-transparent derivation, 
apart from lexical specification as something special. They are consequences of the nature of categories and radical lexicalization.

Also because of the properties described in this section, a string as a category cannot be empty, which would
violate CCG's principle of adjacency and principle of transparency (see Steedman references). No rule
in\xref{rul:fa} or\xref{rul:cs} can apply if one of the categories is empty.
Therefore the surface string  itself for the singleton (\emph{s} in example\xref{ex:singletonapp}) cannot be empty either.

Having explored the possibilities for the singleton types in combinatory categories, we look at their use.

\subsection{Verb-particles and Phrasal Idioms with Singleton Types}\label{sec:analysis}
In verb-particle constructions, the differences in the syntax-semantics correspondence force
the following lexical distinctions. We now write the categories in more detail than in the preview.
\pex \label{ex:pickup}
\a picked := \cgf{(S\bs NP)\fs``up''\fs\cgs{NP}{-heavy}} $:\lambda y\lambda x\lambda z.\so{cause}(\so{init}(\so{hold$_x$}yz))z$
\a picked := \cgf{(S\bs NP)\fs\cgs{NP}{+lexc}\fs``up''}$:\lambda x\lambda y\lambda z.\so{pick$_x$}yz$
\a picked := \cgf{(S\bs NP)\fs NP} $:\lambda x\lambda y.\so{pick}xy \wedge \so{choose}xy$
\xe

The features above are all finite-state computable, just like morphological ones, 
as phonological weight ($\mp$heavy) and lexical content ($\mp$lexc) in an expression substituting for a category. 
All CCG category features can be interpreted this way, because combinators do all the syntactic work.

The reason for having two different grammar entries (a--b) for \emph{pick up} follows from the fact that they are not equally substitutable, for example as an answer to \emph{What did you do?}

\xreff{ex:pickup}{b} leads to achievement, and \xreff{ex:pickup}{a} to culmination. Both cases also differ from (c), which provides wider substitution for \cgf{NP}, and with a different meaning. 
We treat (a--c) 
distinctions surface-compositionally, which are transparently projected without wrap:
\ex \label{ex:pickder}
\footnotesize
\begin{ccg}{4}
{I & picked & the book & up}
{\cat{\cgs{NP}{1s}}&\cat{(S\bs NP)\fs``up"\fs\cgs{NP}{-heavy}} &
\cat{NP} & \cat{((S\bs NP)\bs(S\bs NP))\fs NP}\\
\lf{\so{i\,}} & \lf{\lambda y\lambda x\lambda z.\so{cause}(\so{init}(\so{hold\,}_xyz))z}&
\lf{\so{def\,}\so{book}} & \lf{\lambda x\lambda p\lambda y.\so{up}(py)x}\\
& \cgline{2}{\cgfa}\\
& \cgres{2}{\cat{(S\bs NP)\fs``up"}}\\
&\cgres{2}{\lf{\lambda x\lambda z.\so{cause}(\so{init}(\so{hold\,}_x(\so{def\,}\so{book})z))z}}\\
&\cgline{3}{\cgfa}\\ 
&\cgres{3}{\cat{S\bs NP}}\\
&\cgres{3}{\lf{\lambda z.\so{cause}(\so{init}(\so{hold\,}_{\lambda x\lambda p\lambda y.\so{up}(py)x}}(\so{def\,}\so{book})z))z}\\
\cgline{4}{\cgba}\\ 
\cgres{4}{\cat{S}}\\
\cgres{4}{\lf{\so{cause}(\so{init}(\so{hold\,}_{\lambda x\lambda p\lambda y.\so{up}(py)x}}(\so{def}\so{book})\so{i\,}))\so{i\,}}\\
}
\end{ccg}
\xe
where \so{hold\,} at the end of the derivation can interpret its event modality (contingency) compositionally, since 
it is a closed lambda term.

Notice that the word \emph{up} knows nothing about the verb-particle construction. Its category
is for a PP head, say \cgs{PP}{up}, as a predicate modifier. It is the verb that delivers the distinct meaning. Its subcategorization is for a singleton, which eschews the syntactic category of the word \emph{up}
but not its phonology and semantics, as described in\xreff{ex:singletonapp}{b}.

\xreff{ex:pickup}{b} can be assumed to arise from the syntactic category \cat{VP\fs\cgs{NP}{+lexc}\fs``up''} by finite inflection. CCG has options here, to accommodate morphology
without having to have a ``morphological insertion point'' in a contiguous but MWE entry
\emph{pick up} := \cat{VP\fs\cgs{NP}{+lexc}}, to avoid ?\emph{pick upped}.\footnote{The fact
that this form is also attested in child and adult language suggests that these entries may be bonafide
lexical options.} This is made possible by singleton types.

Examples\xreff{ex:pickup}{a--b} use a degree of freedom which is relevant to phrasal idioms. 
The singleton syntactic type ``up'' corresponding to the LF placeholder $x$ maintains the compositionality
of the correspondence; but, it may have no contribution to the predicate-argument structure at all in some cases, 
which would make it paracompositional, 
because its semantic type is a closed lambda term
as far as predicate-argument structure is concerned. Notice that in\xreff{ex:singletonapp}{b}, \so{s}
is not in the predicate-argument structure of $p$; it is a contingency of $p$.

Consider the following examples in this regard, where $x$ for \so{bucket\,} as an event modality might mean
`ceremonial death', `reported death', etc.:
\pex \label{ex:bucket}
\a kicked := \cgf{(S\bs NP)\fstars``the bucket''}$:\lambda x\lambda y.\so{die}_xy$
\a kicked := \cgf{(S\bs NP)\fs NP}$:\lambda x\lambda y.\so{kick}xy$ 
\xe

They anticipate very limited  syntax in the semantically paracompositional part in the idiom reading (a) because of having to enumerate them (\emph{kick the old/proverbial bucket} vs \emph{kick the bucket that John thought overflowed}).\footnote{It is tempting to try \cgs{NP}{proverbial bucket}\lf{\so{proverb}\so{death}} for \emph{kick the proverbial bucket} which is a head-subcategorizing category; but, we would have to overextend ourselves to eliminate the idiom reading in
\emph{kick the proverbial bucket that overflowed} if we have to. In this sense we suggest that
phrasal idioms are best treated with singleton types.}
These assumptions cannot give rise to
the idiom reading in \emph{the bucket that you kicked}, with no further stipulation than singleton categories in a lexical entry (cf. a--b; '*' on the right of a derivation means it is not possible):
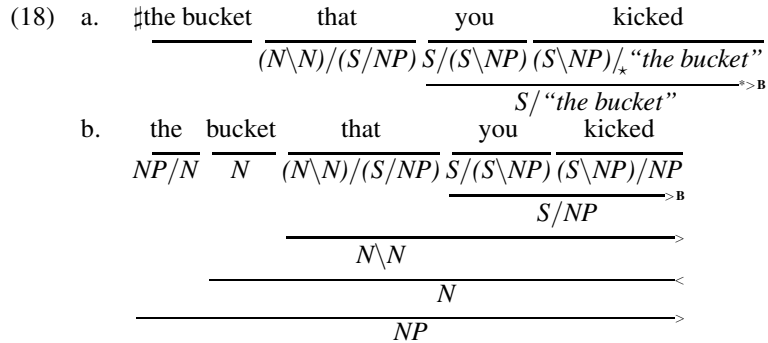
\pex\label{ex:kicks}
\a \begin{ccg}{4}{$\sharp$the bucket & that & you & kicked}
{
 & \cat{(N\bs N)\fs(S\fs NP)} &\cat{S\fs(S\bs NP)} & \cat{(S\bs NP)\fstars``the bucket''}\\
&&\cgline{2}{*\cgfc}\\ &&\cgres{2}{\cat{S\fs``the bucket''}}\\
}
\end{ccg}
\a \begin{ccg}{5}{the & bucket & that & you & kicked}
{
\cat{NP\fs N} & \cat{N} & \cat{(N\bs N)\fs(S\fs NP)} & \cat{S\fs(S\bs NP)} & \cat{(S\bs NP)\fs NP}\\
&&&\cgline{2}{\cgfc}\\ &&&\cgres{2}{\cat{S\fs NP}}\\
&&\cgline{3}{\cgfa}\\ &\cgres{3}{\cat{N\bs N}}\\
& \cgline{4}{\cgba}\\ &\cgres{4}{N}\\
\cgline{5}{\cgfa}\\ \cgres{5}{NP}
}
\end{ccg}
\xe
Given the polyvalent argument category of the relative pronoun, we can see that relativization out of
phrasal idioms would not be possible even if we allowed composition of singleton types, therefore
the syntactic productivity of idiomatically combining phrases arises from their use of head-dependencies rather than singletons, as we shall soon see in derivations similar to (b), in\xref{ex:beans2}.

We note that carrying the head-word in a polyvalent category to have the same effect, for example
\emph{kick} := \cat{(S\bs NP)\fstars\cgs{NP}{bucket}}, overgenerates the idiom reading, because \emph{the bucket that John thought overflowed} can substitute for \cgs{NP}{bucket}.

The direct approach to categories that we see in radically lexicalized grammars, whether they are polyvalently substitutable or not, contrasts
with systems of rewrite and/or record keeping in which post-processing is possible. For example there is no reanalysis or post-processing mechanism needed to eliminate the
idiomatic reading below:
 \ex
\label{ex:drag-rnr} 
 \cgex{6}
  {$\sharp$Mary & dragged & and & John & kicked & the bucket.\\
  \cgul & \cgul &&\cgul & \cgul\\
  \cat{S\fs(S\bs NP)} & \cat{(S\bs NP)\fs NP}
  && \cat{S\fs(S\bs NP)} & \cat{(S\bs NP)\fstars``the bucket''}\\
  \cgline{2}{\cgfc}&&\cgline{2}{*\cgfc}\\
  }
 \xe

We can then follow \cite{steedmanbaldridge-guide} in assuming that passive is a polyvalent  lexical process
headed by the passive morpheme, mapping for example \cat{\cgs{VP}{inf}\fs NP} to \cat{\cgs{VP}{pass}},
which eliminates passivization *\emph{the breeze was shot} from the entry:
\ex
shoot :=\cat{\cgs{VP}{inf}\fs``the breeze''}\lf{\lambda x\lambda y.\so{smalltalk\,}_x\so{one}y}
\xe

Idioms such as \emph{at any rate, beside the point} further demonstrate
that all idioms needing restricted types must contain a predicative element in the domain of locality of their head because we are
required by paracompositionality to record the special reading and contingency, for example
as extension of discursive clarification (a) and  comparison (b):
\pex
\a at := \cat{(S\fs S)\fs``any rate''} \lf{\lambda x\lambda s.\so{more}\so{exactly}s_x}
\a at := \cat{(S\bs S)\fs``any rate''} \lf{\lambda x\lambda s.\so{contrastwith}_x s}

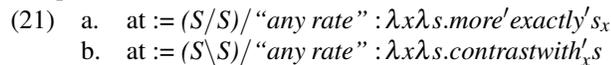
\xe
 
\section{Head-word Subcategorization and Idioms}\label{sec:icp}

The difference between idiomatically combining phrases
and phrasal idioms
such as kicking the bucket is clear: The syntactically active ones are active because the idiomatic part has a role in the predicate-argument structure. `Secret' is an argument of
`divulge', whereas `bucket' is not an argument of 'die'.
For example, \emph{spill the beans} seems to require categorization such as (a) below
in the manner of\xref{ex:kick0-b}, rather than (b) fashioned from\xref{ex:kick0} or singleton-subcategorizing (c).
Cf. also the non-idiomatic \emph{spill} in (d). Tense morphology renders finite versions of  
\cgs{VP}{inf} below as \cat{S\bs NP}, eg. \emph{spilled}:=\cat{(S\bs NP)\fs\cgs{NP}{beans}}
for (a).
\pex\label{ex:spills}
\a spill := \cat{\cgs{VP}{inf}\fs\cgs{NP}{beans}}\lf{\lambda x\lambda y.\so{divulge}_x\so{secret\,}y}
\a spill := \cat{\cgs{VP}{inf}\fs NP}
\lf{\lambda x\lambda y.\mathrm{if\,} {head}(x)=\so{beans} \mathrm
{then\,} \so{divulge}_x\so{secret\,}y\\ 
\hspace*{11em}\mathrm{else\,} \so{spill\,}xy}
\a spill := \cat{(\cgs{VP}{inf}\fs``beans'')\fs PredP}\lf{\lambda p\lambda x\lambda y.\so{divulge}_{px}\so{secret\,}y}
\a spill := \cat{\cgs{VP}{inf}\fs NP}\lf{\lambda x\lambda y.\so{spill\,}xy}
\xe
\cat{PredP} is a predicative phrase type, which includes the quantifier phrase.
The syntactic type of the idiomatic argument in (a) encodes the head-dependency from surface structure. 
It avoids the idiomatic reading in \emph{to spill the bean}, which (b) may not.
(b)-style solutions would depend on LF objects, which may not always reflect surface forms in full.
In fact (b) requires post-processing to eliminate the idiom reading
in the following example:
\ex \label{ex:posthoc}
\small
\hspace*{-2.8em}\begin{ccg}{6}{$\sharp$You & spilled & and & Mary & cooked & the beans}
{
\cat{S\fs(S\bs NP)}
&\cat{(S\bs NP)\fs\cgs{NP}{}}
& \cat{(X\bstars X)\fstars X}
& \cat{S\fs(S\bs NP)}
& \cat{(S\bs NP)\fs NP}
& \cgs{NP}{beans}\\
\lf{\lambda p.p\,\so{you}}&
\lf{\lambda x\lambda y.\mathrm{if\,}\cdots}&
\lf{\lambda p\lambda q\lambda z.\so{and\,}(pz)(qz)}&
\lf{\lambda p.p\,\so{m}}&
\lf{\lambda x\lambda y.\so{cook}xy}&
\lf{\so{def}\so{beans}}
\\
\cgline{2}{\cgfc}  &&\cgline{2}{\cgfc}\\
\cgres{2}{S\fs\cgs{NP}{}} && \cgres{2}{S\fs NP}\\
\cgres{2}{\lf{\lambda x.\mathrm{if\,} head(x)=\cdots}}
&&\cgres{2}{\lf{\lambda x.\so{cook}x\so{m}}}\\
\cgline{5}{\&}\\
\cgres{5}{\cat{S\fs NP}}\\
\cgres{5}{\lf{\lambda z.\so{and\,}(\mathrm{if\,} head(z)=\cdots)(\so{cook}z\,\so{m})}}\\
\cgline{6}{\cgfa}\\ \cgres{6}{\cat{S\fs NP}\lf{\so{and\,}(\mathrm{if\,} head(\so{def}\so{beans})=}}\\
\cgres{6}{$\so{beans}\,\mathrm{then\,}\so{divulge}\so{secret}
\so{you}\cdots)(\so{cook} (\so{def}\so{beans})\,\so{m})$}
} 
\end{ccg}
\xe
This is still the case if we treat the construction as multi-headed, as \cite{gazdar85}:238 do, by also assuming  \emph{the beans} := \cgs{NP}{beans}\lf{\so{secret\,}}, and changing the LF choice condition of \emph{spill} to `if $head(x)$=\so{secret\,} then $\so{divulge}xy$ else $\so{spill\,}xy$'. \so{Cook}
does not refer to this entry.

The process of marking head-word dependencies requires statistical learning, as the
category such as \cgs{NP}{beans} in\xreff{ex:spills}{a} implies. It has been known in TAG systems with supertags 
since \cite{chen:99} that
disambiguating such categories is feasible with training. The earliest approach to such marking
in CCG is \cite{clar:07,clar:02a}  as far as we know, where probabilistic CCGs are similarly trained. 
Later work such as \cite{artz:15} shows further progress in disambiguation of head-dependencies.

\cgs{NP}{beans} is a polyvalent type, not a singleton. Therefore we get the following accounted for 
by\xreff{ex:spills}{a} (some of the examples are from \cite{vill:04}):
\pex \label{ex:dahbeans}
\a spill /several/the musical/the artistic/mountains of/loads of/ beans 
\a spill the beans no one cares about
\xe

Head-marking of an argument category by the idiom's head is required because of examples such as below, where
an idiomatic reading is eliminated despite relatively free syntax because the coordinands would
not be like-typed:
\ex\label{ex:rnr-bad}
\hspace*{-1em}\begin{ccg}{6}{$\sharp$You & spilled & and & Mary & cooked & the beans}
{
\cat{S\fs(S\bs NP)}
&\cat{(S\bs NP)\fs\cgs{NP}{beans}}
& \cat{(X\bstars X)\fstars X}
& \cat{S\fs(S\bs NP)}
& \cat{(S\bs NP)\fs NP}
& \cgs{NP}{beans}\\
\cgline{2}{\cgfc}  &&\cgline{2}{\cgfc}\\
\cgres{2}{S\fs\cgs{NP}{beans}} && \cgres{2}{S\fs NP}\\
\cgline{5}{*\&}
} 
\end{ccg}
\xe
Right-node raising succeeds when non-idiomatic entries such as\xreff{ex:spills}{d} do not subcategorize for head-word marked arguments.\xref{ex:rnr-bad} is unproblematic with it.

When the head of the construction does not require identical types as does
the conjunction above, head-projection works with simple term match; cf. the one for kicking the bucket in\xreff{ex:kicks}{a} ($h$ is for head-word feature):

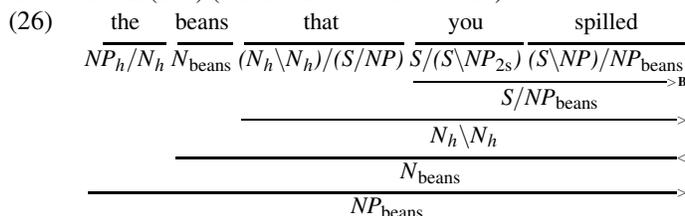
\ex\label{ex:beans2}\small
\begin{ccg}{5}{the & beans & that& you & spilled}
{
\cat{\cgs{NP}{$h$}\fs\cgs{N}{$h$}}&
\cgs{N}{beans} & \cat{(\cgs{N}{$h$}\bs\cgs{N}{$h$})\fs(S\fs NP)}&
\cat{S\fs(S\bs\cgs{\cgs{NP}{2s})}}
&\cat{(S\bs NP)\fs\cgs{NP}{beans}}\\
&&&\cgline{2}{\cgfc}\\
&&&\cgres{2}{S\fs\cgs{NP}{beans}}\\
&& \cgline{3}{\cgfa}\\ &&\cgres{3}{\cgs{N}{$h$}\bs\cgs{N}{$h$}}\\
& \cgline{4}{\cgba}\\&\cgres{4}{\cgs{N}{beans}}\\
\cgline{5}{\cgfa}\\\cgres{5}{\cgs{NP}{beans}}
}
\end{ccg}
\xe
The example also shows that argument types of idiomatically combining phrases must be composable;
therefore; \xreff{ex:spills}{c} is inadequate.\footnote{One way to put it altogether is to use a feature such as $\mp$special in addition to $h$, which ordinary verbs negatively
specify, heads of idiomatic combination positively specify, and heads of syntactic constructions eg. coordinators and relative markers (under)specify as they see fit. The value `+special' need not be further
broken down for singletons because they are self-representing, and, presumably, featureless.
For example phonological weight is intrinsically captured in \cat{``the beans''}; also, lexical
content.}

\section{Idioms Requiring a Combined Approach}\label{sec:combined}

There seems to be cases where a combination of singletons and head-marked polyvalent subcategorization is needed. The \emph{give creeps}
construction, which is sometimes considered not an idiom because of its compositionality \cite{Lars:88}, is paracompositional in our sense, and idiomatically combining in \cite{nunb:94} terminology,
because although \emph{creeps} seems to be an event modality of
\so{revulse} rather than its argument, \so{fear} is an argument. A simple head-marking approach such as `\cat{\fs\cgs{NP}{creeps}}' would overgenerate in cases such
as $\sharp$\emph{give me some creeps}, but we have \emph{give me the
absolute/shivering/full-on creeps}. Notice also that the construction
and related items
 resist dative shift (judgments
are from \cite{lars:12}; `*' seems to be equivalent to
`$\sharp$' in our terms):
\pex
\a The Count gave me the creeps./ *The Count gave the creeps to me.
\a His boss gave Max the boot./ {*}His boss gave the boot to Max.
\xe

Richards \cite{rich:01} observes that (a) below can be the unaccusative of \emph{give}; and, (b) is widely attested in the web (but recall
$\sharp$\emph{give me some creeps}).
\pex
\a Mary got the creeps.
\a give some creeps
\a give := \cat{VP\fs\cgs{N}{creeps}\fs``the''\fs NP}
\small\lf{\lambda x\lambda y\lambda z\lambda w.\so{cause}(\so{init}(\so{revulse}_z\so{fear}_yx))w}
\xe
Assuming that dative shift is polyvalent, following \cite{steedmanbaldridge-guide}, in the form of lexical mapping from \cat{VP\fs NP\fs NP} to \cat{VP\fs\cgs{PP}{to}\fs NP}, we can eliminate it for the type in (c), which
we think captures the insight of Richards, and permits adjunction within an N, e.g. \emph{mountains of creeps}.

Another class of idioms forces a combined approach as well. Semantic reflexives in \emph{I twiddled my thumbs/ate my words/racked my brain/lose my mind} are not morphological reflexives and they are inherently possessive, for example:
\ex
twiddled := \cat{(S\bs\cgs{NP}{agr})\fs``thumbs''\fs\cgs{NP}{-lexc,+poss,agr}}\\ 
\hspace*{12em}\lf{\lambda x\lambda y\lambda z.\so{pass}_y\,\so{time}_{(\so{self\,}z)}z \wedge \so{inalien}(xyz)}
\xe
The LF captures the properties that the subject idles on his own time, the lexical possessive in the LF of $x$ which is presumably lexically \so{poss} is inalienable and belongs to the subject. This is a reflexive in the sense that it must be bound in its local domain
determined by \so{pass.} The referent ($z$) is available in one domain of locality in a radically lexicalized grammar because the head of the idiom does not require a VP in phrase-structure sense but a clause. Agreement is locally available too; by insisting on same agreement features.
The head-dependency is that the argument does not contain lexical material, leaving out examples such as \emph{John twiddled John's thumbs} as an idiom.

\section{No Wrap}\label{sec:nowrap}
We have seen that options\xreff{rul:options}{c/i} and\xreff{rul:options}{c/ii} are not mutually exclusive.
We also suggested that singleton type is a forced move to avoid loss of meaning composition. One consequence of this is the treatment of verb-particles without wrap, which are not related to idioms although they are MWEs.
We now consider option\xreff{rul:options}{b} in more detail from this perspective, which at first sight seems to be just as lexical
as the two alternatives we have considered so far.

The projection principle of CCG, which says that
lexical specification of directionality and order of combination cannot be 
overridden during derivations,
eliminates\xref{rul:badrules} from projection because it
has the second-combining argument (\cat{Y}) of a function applying before its first-combining
argument (\cat{Z}), an operation of the general class that has been proposed
in other categorial approaches under the name of ``wrap.''

\ex \label{rul:badrules}
\begin{tabular}[t]{lllll}
\cgf{(X\fs Y)\fs Z$\colon f$} &Y$\colon a$& $\rightarrow$
& \cgf{X\fs Z$\colon \lambda z.fza$}
\end{tabular}\hfill (*)
\xe

Wrap of the kind in\xref{rul:badrules}{} has a combinatory equivalent, namely Curry's
combinator \combc\,(see \cite{curryfeys58}). CCG's adjacency principle eliminates this combinator on empirical grounds, rather than formal,  as a freely operating rule. Adding\xref{rul:badrules}{} to CCG's projection has the effect of treating VSO and VOS as both grammatical, which is
not the case for Welsh, and to carry the same meaning, which is not the case for Tagalog although both VSO and VOS are fine. These properties must be part of a lexicalized grammar rather than syntactic projection.

The version of wrap which \cite{Bach:76,Dowt:79a,Jaco:92a} employ is different, which was eliminated from consideration so far because it is non-combinatory; and, it violates adjacency of functors and arguments. That wrap is the following:
\ex{\label{rul:surfacewrap}
\cgex{2}{$s_1$ & $s_2$\\
\cgul & \cgul\\
\cgf{X\fs$_W$Y}$:f$ & \cgf{Y}$:a$\\
\cgline{2}{\,wrap}\\
\cgres{2}{first($s_1$)$\,s_2\,$rest($s_1$) := \cgf{X}$:fa$}
}
\xe
where $first()$ function gives the first element in a list of surface expressions for Bach \cite{Bach:76}, or first word for Dowty \cite{Dowt:79a};
and, $rest()$ returns the rest of the expression. The wrapping slash
`\cat{\fs$_W\!$}' of Jacobson \cite{Jaco:92a} does the infixation of $s_2$.}

Semantically, it is function application. Syntactically, no combinator can do what this rule does to its input expressions, which is to rip apart one surface expression ($s_1$) and insert into it. It differs from~\combc, which wraps one independent expression in \emph{two} independent
expressions.

The appeal of surface wrap to MWEs was to be able to write a category for \emph{pick $\cdots$ up}
as for example \emph{pick} := \cat{(S\bs NP)\fs$_W$NP\fs\cgs{P}{up}}\lf{\lambda x\lambda y\lambda z.\so{pick$_x$}yz}; cf.\xref{ex:pickder}.

Syntactic wraps such as above, whether combinatory or non-combinatory,
have domino effects on dependency and constituency, unlike `lexical wrap',
where a lexical entry specifies its correspondence;
for example, for the strictly VSO Welsh verb \emph{gwelodd} := \cat{(S\fs NP)\fs\cgs{NP}{3s}}\lf{\lambda x\lambda y.\so{saw}yx}; note the LF.

An example of global complications in grammar caused by wrap can be seen below, where dashed boxes
denote wrapped-in material; cf. Figure~\ref{fig:subs}.

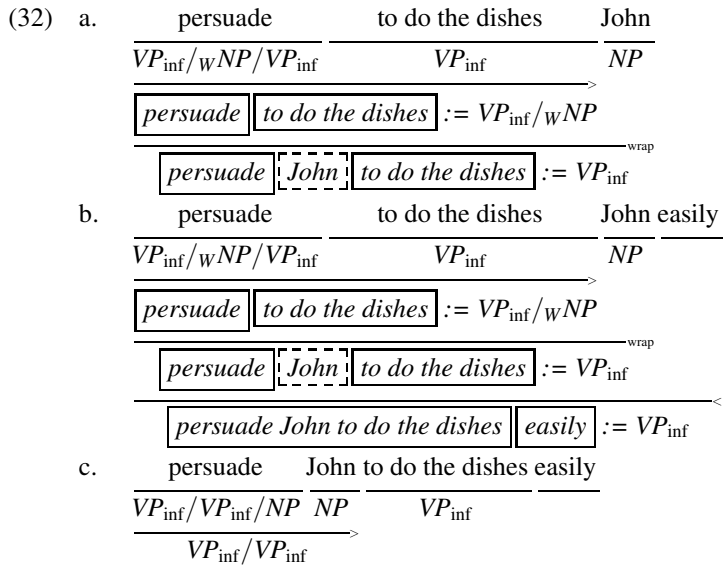
\pex 
\a
\cgex{3}{persuade & to do the dishes & John\\
\cgul & \cgul & \cgul\\
\cgf{\cgs{VP}{inf}\fs$_W$NP\fs\cgs{VP}{inf}} & \cgs{VP}{inf} & \cgf{NP}\\
\cgline{2}{\cgfa}\\[.4ex]
\cgres{2}{\mbox{\fbox{persuade}\,\fbox{to do the dishes}} := \cgs{VP}{inf}\fs$_W$NP}\\
\cgline{3}{\,wrap}\\
\cgres{3}{\mbox{\fbox{persuade}\,\dbox{John}\,\fbox{to do the dishes}} := \cgs{VP}{inf}}
}\vskip .5ex
\a
\cgex{4}{persuade & to do the dishes & John & easily\\
\cgul & \cgul & \cgul&\cgul\\
\cgf{\cgs{VP}{inf}\fs$_W$NP\fs\cgs{VP}{inf}} & \cgs{VP}{inf} & \cgf{NP}\\
\cgline{2}{\cgfa}\\[.4ex]
\cgres{2}{\mbox{\fbox{persuade}\,\fbox{to do the dishes}} := \cgs{VP}{inf}\fs$_W$NP}\\
\cgline{3}{\,wrap}\\
\cgres{3}{\mbox{\fbox{persuade}\,\dbox{John}\,\fbox{to do the dishes}} := \cgs{VP}{inf}}\\
\cgline{4}{\cgba}\\
\cgres{4}{\mbox{\fbox{persuade John to do the dishes}\,\fbox{easily}} := \cgs{VP}{inf}}
}\vskip .5ex
\a
\cgex{4}{persuade & John & to do the dishes & easily\\
\cgul & \cgul & \cgul&\cgul\\
\cgf{\cgs{VP}{inf}\fs\cgs{VP}{inf}\fs NP}&\cgf{NP} & \cgs{VP}{inf}\\
\cgline{2}{\cgfa}\\
\cgres{2}{\cgs{VP}{inf}\fs\cgs{VP}{inf}}\\
}
\xe
Derivation (a) is Bach's use of non-combinatory wrap rule in\xref{rul:surfacewrap}. Given these categories
which involve wrap, there is one interpretation for (b), where the adverb can only modify \emph{persuade}. With the unwrapped version of \emph{persuade} in (c), two
interpretations are possible: one modifies the VP complement of \emph{persuade}, and the other, \emph{persuade John}, both of which are required for adequacy.


\section{Conclusion}\label{sec:conc}
One point of departure of CCG from other categorial grammars and from tree-rewrite systems is that (i) we can complicate the basic vocabulary of the theory, but (ii) not its basic mechanism such as introducing wrap, if a better explanation can be achieved. The  first point has been made by Chomsky repeatedly since \cite{chomsky72}:68.
Singleton types could be viewed as one way of doing that, much like \cgf{S\bs NP} vs. \cgf{VP} distinction. We have argued that it is actually not a complication at all in CCG's case, because the possibility has been available, in the notion of type as a set of values, which can be a singleton set. CCG differs from Chomskyan notion of category substitution by
eliminating \emph{move}, empty categories and lexical insertion altogether, which means that all computation is local, type-driven, and there is no action-at-a-distance, to address the second point. The expressions substituting for these types are then locally available in the course of a derivation. This seems critical for MWEs.

The possibility of a singleton value is built-in to any type. The asymmetry
of CCG's singletons' categorization, that they can be arguments, and arguments of arguments and results,
and, their inherent applicative nature,
deliver MWEs and phrasal idioms as natural consequences rather than stipulation or a ``pain in the neck for NLP.''
Syntactically active idioms are not singleton-typed because they have
relevance to predicate-argument structure; and, their narrower syntax, compared to free syntax, seems to necessitate head-marking of some argument categories, which is known to be probabilistically learnable.

Some implications of our analyses are that all idioms  can be made compositional at the level of
a lexical correspondence without losing semantic distinctions, and without meaning postulates
or reanalysis.
Categorial post-processing of MWEs and phrasal idioms,
and multi-stage processing of them in the lexicon, as done by \cite{cope:94,vill:04}, may be unnecessary 
if we assume type substitution to be potentially having one value, and surface head-marking to be an option for
polyvalent argument types. One conjecture is that any idiom
in any language has to involve a predicate implicated by some predicative element in the expression to keep the meaning assembly paracompositional.

The analyses in the article can be replicated by running the CCG tool at \url{github.com/bozsahin/ccglab}.
The particular fragment in the chapter is at \url{github.com/bozsahin/ccglab-grammars/cb-ag-fg2018-grammar.}
\small
\bibliography{cem,mark-references}

\begin{thebibliography}{33}
\providecommand{\natexlab}[1]{#1}
\providecommand{\url}[1]{\texttt{#1}}
\providecommand{\urlprefix}{}

\bibitem[{Artzi et~al.(2015)Artzi, Lee, and Zettlemoyer}]{artz:15}
Artzi, Y., Lee, K., Zettlemoyer, L.: Broad-coverage {CCG} semantic parsing with
  {AMR}.
\newblock In: Proceedings of the 2015 Conference on Empirical Methods in
  Natural Language Processing. pp. 1699--1710. Lisbon, Portugal (2015)

\bibitem[{Bach(1976)}]{Bach:76}
Bach, E.: An extension of {C}lassical {T}ransformational grammar.
\newblock In: Problems in Linguistic Metatheory: Proceedings of the 1976
  Conference at Michigan State University. pp. 183--224. Michigan State
  University, Lansing, MI (1976)

\bibitem[{Baldridge and Kruijff(2003)}]{Bald:03}
Baldridge, J., Kruijff, G.J.: Multi-modal {C}ombinatory {C}ategorial {G}rammar.
\newblock In: Proceedings of 11th Annual Meeting of the European Association
  for Computational Linguistics. pp. 211--218. Budapest (2003)

\bibitem[{Bond et~al.(2015)Bond, Ho, and Flickinger}]{bond:15}
Bond, F., Ho, J.Q., Flickinger, D.: Feeling our way to an analysis of english
  possessed idioms.
\newblock In: M{\"u}ller, S. (ed.) Proceedings of the 22nd International
  Conference on {H}ead-{D}riven {P}hrase {S}tructure {Gr}ammar. pp. 61--74.
  CSLI Publications, Stanford, CA (2015)

\bibitem[{Boz\c{s}ahin(2012)}]{bozs:12}
Boz\c{s}ahin, C.: Combinatory Linguistics.
\newblock De Gruyter Mouton, Berlin (2012)

\bibitem[{Chen et~al.(1999)Chen, Bangalore, and Vijay-Shanker}]{chen:99}
Chen, J., Bangalore, S., Vijay-Shanker, K.: New models for improving supertag
  disambiguation.
\newblock In: Proceedings of the ninth conference on European chapter of the
  Association for Computational Linguistics. pp. 188--195. Association for
  Computational Linguistics (1999)

\bibitem[{Chomsky(1972)}]{chomsky72}
Chomsky, N.: Some empirical issues in the theory of transformational grammar.
\newblock In: Peters, S. (ed.) Goals of Linguistic Theory. Prentice-Hall,
  Englewood Cliffs, NJ (1972)

\bibitem[{Clark and Curran(2007)}]{clar:07}
Clark, S., Curran, J.R.: Wide-coverage efficient statistical parsing with {CCG}
  and log-linear models.
\newblock Computational Linguistics 33(4), 493--552 (2007)

\bibitem[{Clark et~al.(2002)Clark, Hockenmaier, and Steedman}]{clar:02a}
Clark, S., Hockenmaier, J., Steedman, M.: Building deep dependency structures
  with a wide-coverage {CCG} parser.
\newblock In: Proceedings of the 40th Annual Meeting on Association for
  Computational Linguistics. pp. 327--334 (2002)

\bibitem[{Copestake(1994)}]{cope:94}
Copestake, A.: Representing idioms.
\newblock In: {HPSG} Conference. Copenhagen (1994)

\bibitem[{Curry and Feys(1958)}]{curryfeys58}
Curry, H.B., Feys, R.: Combinatory Logic.
\newblock North-Holland, Amsterdam (1958)

\bibitem[{Dowty(1979)}]{Dowt:79a}
Dowty, D.: Dative movement and {T}homason's extensions of {M}ontague {G}rammar.
\newblock In: Davis, S., Mithun, M. (eds.) Linguistics, Philosophy, and
  Montague Grammar, pp. 153--222. University of Texas Press, Austin (1979)

\bibitem[{Dowty(2003)}]{dowt:03}
Dowty, D.: The dual analysis of adjuncts/complements in categorial grammar.
\newblock pp. 33--66 (2003), in \cite{lang:03}

\bibitem[{Everaert et~al.(1995)Everaert, Van~der Linden, Schreuder, and
  Schenk}]{ever:95}
Everaert, M., Van~der Linden, E.J., Schreuder, R., Schenk, A. (eds.): Idioms:
  structural and psychological perspectives.
\newblock Lawrence Erlbaum, New Jersey (1995)

\bibitem[{Gazdar et~al.(1985)Gazdar, Klein, Pullum, and Sag}]{gazdar85}
Gazdar, G., Klein, E., Pullum, G., Sag, I.: Generalized Phrase Structure
  Grammar.
\newblock Harvard University Press (1985)

\bibitem[{Jacobson(1992)}]{Jaco:92a}
Jacobson, P.: Flexible categorial grammars: Questions and prospects.
\newblock In: Levine, R. (ed.) Formal Grammar, pp. 129--167. Oxford University
  Press, Oxford (1992)

\bibitem[{Kay et~al.(ms.)Kay, Sag, and Flickinger}]{kay:idioms}
Kay, P., Sag, I.A., Flickinger, D.: A lexical theory of phrasal idioms (ms),
  available at: www1.icsi.berkeley.edu/$\sim$kay/idiom-pdflatex.11-13-15.pdf

\bibitem[{Lang et~al.(2003)Lang, Maienborn, and Fabricius-Hansen}]{lang:03}
Lang, E., Maienborn, C., Fabricius-Hansen, C.: Modifying adjuncts.
\newblock Walter de Gruyter (2003)

\bibitem[{Larson(1988)}]{Lars:88}
Larson, R.: On the double object construction.
\newblock Linguistic Inquiry 19, 335--392 (1988)

\bibitem[{Larson(2012)}]{lars:12}
Larson, R.: On ``dative idioms'' in {E}nglish.
\newblock In: Workshop on Syntax-semantics. Fuji Women's University (2012)

\bibitem[{Lichte and Kallmeyer(2016)}]{lich:16}
Lichte, T., Kallmeyer, L.: Same syntax, different semantics: A compositional
  approach to idiomaticity in multi-word expressions.
\newblock In: Pi{\~{n}}{\'{o}}n, C. (ed.) Empirical Issues in Syntax and
  Semantics, vol.~11, pp. 111--140. CSSP, Paris (2016)

\bibitem[{van~der Linden(1992)}]{lind:92}
van~der Linden, E.J.: Incremental processing and the hierarchical lexicon.
\newblock Computational linguistics 18(2), 219--238 (1992)

\bibitem[{Manning(1996)}]{manning96}
Manning, C.D.: Ergativity: Argument Structure and Grammatical Relations.
\newblock CSLI, Stanford, CA (1996)

\bibitem[{Moens and Steedman(1988)}]{Moen:88a}
Moens, M., Steedman, M.: Temporal ontology and temporal reference.
\newblock Computational Linguistics 14, 15--28 (1988), reprinted in Inderjeet
  Mani, James Pustejovsky, and Robert Gaizauskas (eds.) {\em The Language of
  Time: A Reader}. Oxford University Press, 93-114.

\bibitem[{Nunberg et~al.(1994)Nunberg, Sag, and Wasow}]{nunb:94}
Nunberg, G., Sag, I.A., Wasow, T.: Idioms.
\newblock Language 70(3), 491--538 (1994)

\bibitem[{Pulman(1993)}]{pulm:93}
Pulman, S.G.: The recognition and interpretation of idioms.
\newblock In: Cacciari, C., Tabossi, P. (eds.) Idioms: Processing, structure,
  and interpretation, pp. 249--270. Lawrence Erlbaum, Hillsdale, NJ (1993)

\bibitem[{Richards(2001)}]{rich:01}
Richards, N.: An idiomatic argument for lexical decomposition.
\newblock Linguistic inquiry 32(1), 183--192 (2001)

\bibitem[{Sag et~al.(2002)Sag, Baldwin, Bond, Copestake, and
  Flickinger}]{sag:02}
Sag, I.A., Baldwin, T., Bond, F., Copestake, A., Flickinger, D.: Multiword
  expressions: A pain in the neck for {NLP}.
\newblock In: International Conference on Intelligent Text Processing and
  Computational Linguistics. pp. 1--15. Mexico City (2002)

\bibitem[{Schuler and Joshi(2011)}]{schu:11a}
Schuler, W., Joshi, A.: Tree-rewriting models of multi-word expressions.
\newblock In: Proceedings of the ACL Workshop on Multiword Expressions: from
  Parsing and Generation to the Real World. pp. 25--30. ACL, Portland, OR
  (2011)

\bibitem[{Steedman(2000)}]{Stee:99}
Steedman, M.: The Syntactic Process.
\newblock MIT Press, Cambridge, MA (2000)

\bibitem[{Steedman(2012)}]{Stee:11a}
Steedman, M.: Taking Scope: The Natural Semantics of Quantifiers.
\newblock MIT Press, Cambridge, MA (2012)

\bibitem[{Steedman and Baldridge(2011)}]{steedmanbaldridge-guide}
Steedman, M., Baldridge, J.: {C}ombinatory {C}ategorial {G}rammar.
\newblock In: Borsley, R., B{\"o}rjars, K. (eds.) Non-transformational syntax,
  pp. 181--224. Blackwell, Oxford (2011)

\bibitem[{Villavicencio et~al.(2004)Villavicencio, Copestake, Waldron, and
  Lambeau}]{vill:04}
Villavicencio, A., Copestake, A., Waldron, B., Lambeau, F.: Lexical encoding of
  {MWE}s.
\newblock In: Proceedings of the Workshop on Multiword Expressions: Integrating
  Processing. pp. 80--87. Association for Computational Linguistics (2004)

\end{thebibliography}

\end{document}